\documentclass[11pt]{article}

\usepackage{acl}
\usepackage{times}
\usepackage{latexsym}
\usepackage[T1]{fontenc}
\usepackage[utf8]{inputenc}
\usepackage{microtype}
\usepackage{inconsolata}
\usepackage{booktabs}
\usepackage{graphicx}
\usepackage{hyperref}
\usepackage{pgfplots}
\pgfplotsset{compat=1.16}

\title{Beyond Helpfulness: A Teaching-over-Solving Diagnostic for Measuring Educational Impact in LLM Tutors}

\author{
Junyi Yao$^{1}$ \and Zihao Zheng$^{1}$ \and Baichuan Li$^{2}$ \\
$^{1}$Washington University in St. Louis, St. Louis, MO, USA \\
$^{2}$Southern Methodist University, Dallas, TX, USA \\
\texttt{j.yao@wustl.edu} \quad \texttt{z.zihaogary@wustl.edu} \quad \texttt{baichuanl@smu.edu}
}

\begin{document}
\maketitle

\begin{abstract}
Large language models are increasingly proposed as educational tutors, yet stronger task-solving ability does not necessarily imply stronger learning support. Motivated by recent calls to measure the social impact of NLP systems in practice, we study whether public LLM tutoring benchmarks distinguish learning-supportive behavior from mere answer production. We propose a lightweight diagnostic based on the gap between solving-oriented and pedagogy-oriented benchmark performance. Using public MathTutorBench leaderboard results, we show that these dimensions are only partially aligned: across eight publicly reported models, the Pearson correlation between solving and pedagogy composites is 0.421, the rank-based Spearman correlation is 0.690, and several models shift meaningfully in rank when evaluation moves from solving to pedagogy. We then analyze the public TutorBench sample and show that agency-relevant behaviors are explicitly encoded in benchmark rubrics, especially in active-learning settings that reward guiding questions, calibrated hints, and non-disclosive scaffolding. Together, these findings suggest that educational-impact evaluation should not treat task success as a sufficient proxy for learning support. We argue that public tutoring benchmarks can better support positive-impact evaluation by reporting solving-oriented and pedagogy-oriented scores separately and by making disclosure-sensitive, student-agency-preserving criteria more explicit.
\end{abstract}

\section{Introduction}

Large language models (LLMs) are increasingly framed as scalable tutors, study partners, and classroom assistants. Yet in education, a response that looks helpful can still be pedagogically harmful: directly giving away an answer may improve immediate task completion while reducing the learner's opportunity to reason independently. This creates a measurement problem for educational NLP. If evaluation centers correctness, fluency, and generic helpfulness alone, it may miss whether a system actually supports learning.

This concern fits a broader question raised in recent NLP-for-positive-impact work: how should we measure the social impact of language technologies beyond narrow model performance \citep{jin-etal-2021-how-good}? In education, the challenge is especially sharp because a tutoring system's impact depends not only on whether it can solve problems, but also on whether it preserves student agency and provides calibrated scaffolding. In that sense, trustworthiness is not only about answer correctness, but also about whether a model can recognize when direct answer delivery would be pedagogically inappropriate. Real-world deployment work such as Tutor CoPilot shows that pedagogical strategies like guiding questions and non-disclosive support matter in practice \citep{wang-etal-2025-tutorcopilot}.

Recent tutoring benchmarks point in the same direction. MathDial explicitly distinguished tutoring from solving \citep{macina-etal-2023-mathdial}, and GuideEval argues that adaptive instructional guidance should be evaluated directly rather than inferred from generic Socratic style \citep{liu-etal-2025-guideeval}. In educational settings, this matters because a model can appear supportive while still over-serving the learner's desire for shortcuts.

This paper makes a compact empirical contribution tailored to this gap and to the constraints of a workshop short paper. Rather than proposing a new benchmark or training a new tutor, we reanalyze public benchmark artifacts to ask a narrower question: do current public tutoring evaluations already separate behaviors that are plausibly connected to learning support from behaviors that merely complete the task for the learner? We do not claim to measure downstream learning gains. Instead, we ask a prior measurement question and answer it with two lightweight studies:

\begin{itemize}
    \item a reaggregation of public MathTutorBench leaderboard scores into solving-oriented and pedagogy-oriented composites; and
    \item a rubric analysis of the public TutorBench sample to inspect which behaviors benchmark designers explicitly reward.
\end{itemize}

Our results are simple but revealing. Solving and pedagogy composites are only moderately correlated ($r=0.421$), and some models shift substantially in rank when evaluation moves from solving to pedagogy. Meanwhile, TutorBench rubrics explicitly encode active-learning behaviors such as giving hints, asking guiding questions, and avoiding premature solution disclosure. Together, these analyses support a practical claim for educational NLP: positive impact should be measured not only by whether an LLM can answer, but also by whether it teaches without taking over the student's reasoning.

More concretely, this paper contributes:
\begin{itemize}
    \item a lightweight \textbf{diagnostic} for separating answer-production ability from learning-support behavior in public tutoring benchmarks;
    \item an empirical demonstration on MathTutorBench that solving and pedagogy composites are only partially aligned;
    \item a rubric-grounded validation from TutorBench that \textbf{student agency} and \textbf{non-disclosive scaffolding} are already encoded in public benchmark criteria; and
    \item a positive-impact measurement template for education-oriented NLP work that lacks private classroom data.
\end{itemize}

\section{Related Work}

MathDial introduced a tutoring dataset grounded in math reasoning and rich pedagogical annotations, showing that tutoring quality cannot be reduced to answer quality alone \citep{macina-etal-2023-mathdial}. More recently, MathTutorBench and GuideEval argued for evaluating open-ended pedagogical capability and adaptive instructional guidance directly rather than inferring them from generic helpfulness \citep{macina-etal-2025-mathtutorbench,liu-etal-2025-guideeval}. The BEA 2025 shared task further operationalized pedagogical dimensions such as mistake identification, guidance quality, and actionability \citep{kochmar-etal-2025-bea-shared-task}. At a broader level, educational NLP has also begun to model classroom discourse moves directly \citep{suresh-etal-2022-fine}. Our paper is narrower than these efforts: we do not propose a new benchmark, but instead extract a compact positive-impact diagnostic from public benchmark artifacts.

\paragraph{Educational impact and deployment.}
Tutor CoPilot provides an important deployment-grounded reference point by demonstrating that AI support for human tutors can improve outcomes in live tutoring while also shifting tutor behavior toward more effective strategies \citep{wang-etal-2025-tutorcopilot}. This paper does not claim comparable real-world evidence. Instead, it contributes a compact benchmark-based analysis intended for early-stage measurement work aligned with positive-impact concerns.

\section{Public Benchmark Reframing}

We use two public sources. The first is the MathTutorBench leaderboard \citep{macina-etal-2025-mathtutorbench}, which reports nine task-level scores for eight tutoring models. The second is the public TutorBench sample, which includes 30 scenarios with structured rubric annotations. Both sources are public and require no student-private data.

The two sources play complementary roles. MathTutorBench provides a model-level view of how current LLM tutors are scored across distinct pedagogical tasks. TutorBench provides a rubric-level view of what benchmark designers explicitly consider good tutoring behavior. Using both lets us ask not only whether tutoring-related dimensions diverge in model performance, but also whether that divergence matters under publicly articulated pedagogical criteria.

For MathTutorBench, we create two composite views:
\begin{itemize}
    \item \textbf{Solving composite}: problem solving, solution correctness, mistake location, and mistake correction.
    \item \textbf{Pedagogy composite}: socratic questioning, scaffolding win rate, pedagogy instruction-following win rate, hard scaffolding, and hard pedagogy instruction-following.
\end{itemize}

Formally, for model $m$, we define:
\begin{itemize}
    \item $\mathrm{Solve}(m)$ as the mean of the four solving-oriented scores,
    \item $\mathrm{Teach}(m)$ as the mean of the five pedagogy-oriented scores, and
    \item $\mathrm{Gap}(m)=\mathrm{Teach}(m)-\mathrm{Solve}(m)$.
\end{itemize}
Positive values indicate models that are relatively stronger on pedagogy than on solving; negative values indicate the reverse.

We treat mistake location and mistake correction as solving-oriented because they primarily assess whether the model can identify or repair the mathematical state of the problem. In contrast, Socratic questioning, scaffolding, and pedagogy instruction following more directly evaluate how the model structures the learner's reasoning process.

For TutorBench, we analyze rubric metadata to identify whether benchmark criteria explicitly reward agency-preserving tutoring behaviors. We use a deterministic rubric-label mapping rather than new human annotation. Rubric items are counted as \textit{agency-dimension} items when their metadata involves instruction following or student-level calibration, and as \textit{agency-skill} items when they involve guiding questions, hinting, eliciting reflection, or learner-state diagnosis. This is not a student-outcome measure; rather, it is an analysis of which pedagogical priorities are operationalized in a public benchmark.

\section{Study 1: Composite Gap Analysis}

Across the eight public MathTutorBench leaderboard models, the Pearson correlation between solving and pedagogy composites is 0.421, while the rank-based Spearman correlation is 0.690. Given the small number of models, we use these values descriptively rather than as hypothesis tests. They suggest that the two capabilities are related but far from interchangeable. A leave-one-out robustness check yields positive Pearson correlations in every seven-model subset (0.251--0.887). Removing Qwen2.5-Math-7B-Instruct increases the Pearson correlation to 0.887, confirming that the public sample is statistically fragile, but the qualitative conclusion remains the same: leaderboard rank can change materially when evaluation shifts from solving to pedagogy.

\begin{table}[t]
\centering
\footnotesize
\setlength{\tabcolsep}{3pt}
\begin{tabular}{lcc}
\toprule
Model & Solving & Pedagogy \\
\midrule
LearnLM-1.5-Pro & 0.750 & 0.594 \\
GPT-4o & 0.695 & 0.592 \\
LLaMA3.1-70B & 0.593 & 0.520 \\
LLaMA3.1-8B & 0.427 & 0.504 \\
LLaMA3.2-3B & 0.453 & 0.482 \\
Llemma-7B & 0.432 & 0.388 \\
Qwen2.5-7B-SocraticLM & 0.350 & 0.332 \\
Qwen2.5-Math-7B & 0.568 & 0.116 \\
\bottomrule
\end{tabular}
\caption{Compact composite scores from public MathTutorBench leaderboard results.}
\label{tab:composite}
\end{table}

\begin{figure}[t]
\centering
\begin{tikzpicture}
\begin{axis}[
    width=\columnwidth,
    height=0.72\columnwidth,
    xmin=0.30, xmax=0.80,
    ymin=0.00, ymax=0.80,
    xlabel={Solving composite},
    ylabel={Pedagogy composite},
    grid=both,
    grid style={line width=.1pt, draw=gray!20},
    major grid style={line width=.2pt,draw=gray!35},
    tick label style={font=\small},
    label style={font=\small},
    legend style={draw=none, font=\scriptsize, at={(0.02,0.98)}, anchor=north west},
]
\addplot[
    only marks,
    mark=*,
    mark size=2pt,
    color=blue
] coordinates {
    (0.750,0.594)
    (0.695,0.592)
    (0.593,0.520)
    (0.427,0.504)
    (0.453,0.482)
    (0.432,0.388)
    (0.350,0.332)
    (0.568,0.116)
};
\addplot[dashed, color=black!45] coordinates {(0.30,0.30) (0.80,0.80)};
\node[anchor=west, font=\scriptsize] at (axis cs:0.750,0.594) {LearnLM-1.5-Pro};
\node[anchor=west, font=\scriptsize] at (axis cs:0.695,0.592) {GPT-4o};
\node[anchor=west, font=\scriptsize] at (axis cs:0.593,0.520) {LLaMA3.1-70B};
\node[anchor=west, font=\scriptsize] at (axis cs:0.427,0.504) {LLaMA3.1-8B};
\node[anchor=west, font=\scriptsize] at (axis cs:0.453,0.482) {LLaMA3.2-3B};
\node[anchor=west, font=\scriptsize] at (axis cs:0.432,0.388) {Llemma-7B};
\node[anchor=west, font=\scriptsize] at (axis cs:0.350,0.332) {Qwen2.5-7B-SocraticLM};
\node[anchor=west, font=\scriptsize] at (axis cs:0.568,0.116) {Qwen2.5-Math-7B};
\end{axis}
\end{tikzpicture}
\caption{Solving versus pedagogy composites. Models below the diagonal are relatively stronger at solving than at pedagogy.}
\label{fig:scatter}
\end{figure}
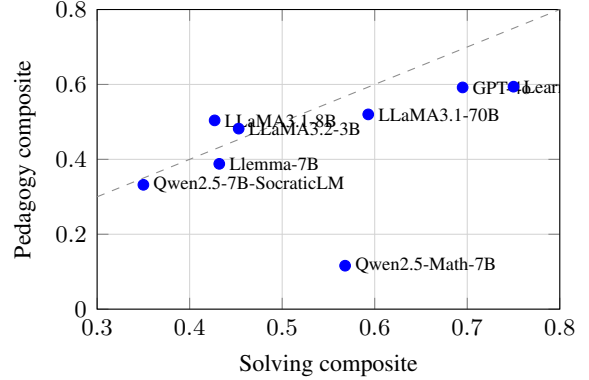

Several models exhibit meaningful divergence once evaluation shifts from solving to pedagogy. LLaMA3.1-8B-Instruct rises from rank 7 on solving to rank 4 on pedagogy, yielding a positive teaching-over-solving gap of $+0.077$. In contrast, Qwen2.5-Math-7B-Instruct ranks 4 on solving but last on pedagogy, with a gap of $-0.452$. This is the clearest example in the public leaderboard of a model that appears comparatively strong if one emphasizes task-solving but weak if one emphasizes teaching-related support.

The moderate correlation is not evidence that solving and pedagogy are unrelated; rather, it shows that solving ability is an insufficient stand-in for pedagogy when the evaluation target is educational impact. A model that answers well but discloses too much, fails to calibrate to confusion, or does not preserve room for student reasoning may still be misaligned with positive educational impact. Benchmark reporting should therefore preserve the distinction between solving strength and teaching strength instead of collapsing them into a single notion of helpfulness.

\section{Study 2: Rubric-grounded Agency Analysis}

To examine whether this interpretation is grounded in benchmark design, we analyze the TutorBench sample rubrics. The sample contains 30 tutoring scenarios and 324 rubric items across three use-case families: adaptive explanation, assessment and feedback, and active learning.

Table~\ref{tab:rubrics} reports a compact summary. We compute agency-dimension density conservatively as the proportion of rubric items containing at least one agency-relevant dimension tag, and agency-skill density as the proportion containing an explicit tutoring-skill tag such as guiding questions or learner-state diagnosis.

\begin{table}[t]
\centering
\small
\setlength{\tabcolsep}{3.5pt}
\begin{tabular}{lccc}
\toprule
Use case & Rubrics & Agency dim. & Agency skill \\
\midrule
Adaptive explanation & 108 & 0.583 & 0.139 \\
Assessment/feedback & 109 & 0.688 & 0.294 \\
Active learning & 107 & 0.720 & 0.411 \\
\bottomrule
\end{tabular}
\caption{Agency-oriented rubric density in the public TutorBench sample.}
\label{tab:rubrics}
\end{table}

Even under this stricter item-level counting rule, the active-learning subset has the highest concentration of agency-oriented criteria. In qualitative examples, rubric language explicitly rewards asking guiding questions, providing hints without disclosing the full next step, and prompting the learner to consider alternative lines of reasoning. One representative criterion requires the tutor to ``provide hint[s] ... by asking at least 2 guiding questions without completely showing next steps.'' This matters because our positive-impact interpretation is not imposed from outside the benchmark; it is already reflected in the public annotation logic. The mapping itself is deterministic and should be read as a transparent heuristic rather than as a validated human-coding study; its value is that the exact rules are inspectable and reproducible.

This point is important for measurement design. If public tutoring rubrics already distinguish between merely continuing the interaction and productively structuring the learner's reasoning, then benchmark reporting should preserve that distinction explicitly. Otherwise, a model can gain credit for sounding supportive while still optimizing for rapid answer delivery. In that sense, the rubric analysis does not only validate our diagnostic post hoc; it also suggests that existing benchmark resources are already rich enough to support more learning-centered leaderboard practices.

\section{Discussion}

Together, the two studies support a compact but actionable conclusion. First, public leaderboard results show that solving strength and pedagogy are only partially aligned. Second, public rubric design shows that benchmark creators already treat agency-preserving guidance as a distinct pedagogical objective. These observations suggest that educational NLP evaluation should foreground learning support as its own measurement target. For a positive-impact workshop setting, the broader implication is that educational benefit should be operationalized through pedagogical support, not inferred from generic utility proxies alone. Read through a trustworthy-AI lens, this is also a decision-control problem under uncertainty: when a tutor model is uncertain about learner state or about the pedagogical appropriateness of intervention, safer behavior may require calibrated restraint rather than maximal answer delivery.

\paragraph{Implications for benchmark reporting.}
We recommend that tutoring benchmark leaderboards report solving-oriented and pedagogy-oriented scores separately, rather than only aggregate helpfulness or overall quality. At minimum, tutoring benchmarks should include a disclosure-sensitive dimension that penalizes giving away full solutions when the learner context calls for scaffolding.

This paper is intentionally lightweight. Its contribution is methodological: a short-paper demonstration that public educational benchmarks already contain signals relevant to social impact, provided that we aggregate and interpret them appropriately.

\section{Conclusion}

If LLM tutors are to be evaluated through the lens of positive impact, then generic helpfulness is not enough. Public tutoring benchmarks already show that models can be stronger at solving than at teaching, and that benchmark rubrics explicitly value scaffolding and student agency. A compact teaching-over-solving analysis offers one practical step toward learning-centered evaluation in educational NLP, especially for short-form work that aims to connect benchmark evidence to social impact without overclaiming deployment or learning effects.

\section{Limitations}

Our analyses do not measure downstream student outcomes, so they should not be interpreted as evidence that one model causes better learning than another. The MathTutorBench reanalysis is also based on only eight public models, which makes the reported Pearson correlation statistically fragile; we therefore present it as a descriptive diagnostic and report leave-one-out robustness rather than a population estimate. Our claims also depend on benchmark design choices: different task groupings or alternative rubrics could change the reported gaps. Alternative task groupings are possible; our grouping is intended as a transparent diagnostic rather than a definitive taxonomy. In addition, the TutorBench analysis uses a 30-example public sample and a deterministic metadata mapping rather than independent human coders, so it should be read as an auditable heuristic characterization of public rubric priorities, not as a validated annotation study. Finally, because we analyze public artifacts rather than raw generations, we do not directly measure answer disclosure turn by turn.

\section{Ethical Considerations}

This study uses only public benchmark artifacts and no private student data. We stress, however, that benchmark-aligned pedagogical behavior should not be conflated with demonstrated learning gains, and that overreading benchmark scores as deployment readiness could encourage harmful decisions or metric gaming \citep{yao-etal-2026-ranking-abuse,sang2026uncertaintyawarerobust,zheng2026reliablefinancialnamedentity}. Related cross-domain work on privacy-preserving learning under heterogeneous deployment constraints likewise shows that responsible evaluation must account for operational conditions rather than idealized performance alone \citep{hou-etal-2026-sar-rafl}. In that sense, trustworthy tutor evaluation should reflect internal decision quality, not just surface compliance, and may sometimes favor withholding a direct solution when answer-giving is not pedagogically justified \citep{chen2026beyondexternalconstraints}.

\bibliography{references}

\end{document}